\documentclass[letterpaper, 10 pt, conference]{ieeeconf}
\usepackage[utf8]{inputenc}
\usepackage{amsmath}
\usepackage{amssymb}
\usepackage{graphicx}
\usepackage{xcolor}
\usepackage{subcaption}

\IEEEoverridecommandlockouts                              

\title{\LARGE \bf
Towards biomimicry of a bat-style perching maneuver on structures: the manipulation of inertial dynamics
}

\author{Alireza Ramezani
\thanks{Alireza Ramezai is with the Department of Electrical and Computer Engineering, Northeastern University, Boston, MA 02115
        {\tt\small a.ramezani@northeastern.edu}}%
}

\begin{document}


\maketitle

\begin{abstract}
The flight characteristics of bats remarkably have been overlooked in aerial drone designs. Unlike other animals, bats leverage the manipulation of inertial dynamics to exhibit aerial flip turns when they perch. Inspired by this unique maneuver, this work develops and uses a tiny robot called \textit{Harpoon} to demonstrate that the preparation for upside-down landing is possible through: 1) reorientation towards the landing surface through zero-angular-momentum turns and 2) reaching to the surface through shooting a detachable landing gear. The closed-loop manipulations of inertial dynamics takes place based on a symplectic description of the dynamical system (body and appendage), which is known to exhibit an excellent geometric conservation properties. 
\end{abstract}


%
\IEEEpeerreviewmaketitle

\section{Introduction}
The key hypothesis this work tries to inspect is: ''The feasibility of performing flip turns, a well known attribute of a bat-like landing maneuver, through the manipulation of inertial dynamics while excluding aerodynamic forces.'' Bats (and birds) possess no energy-hungry motors widely used in flying robots for thrust vectoring yet they are more capable than any of these systems when agility and, of course, energy efficiency of flight are concerned. 



Flying vertebrates apply the combination of inertial dynamics and aerodynamics manipulations to showcase extremely agile maneuvers. Unlike rotary- and fixed-wing  systems wherein aerodynamic surface (i.e., ailerons, rudders, propellers, etc.) come with the sole purpose of aerodynamic force adjustment, the wings (also called appendages) in birds and bats possess more sophisticated roles. It is known that birds perform zero-angular-momentum turns by making differential adjustments (e.g., collapsing armwings) in the inertial forces led by one wing versus the other. Or, bats apply a similar mechanism to perform sharp banking turns \cite{riskin2012upstroke}.


\begin{figure}[!t]
	\centering
		\includegraphics[width=0.3\textwidth]{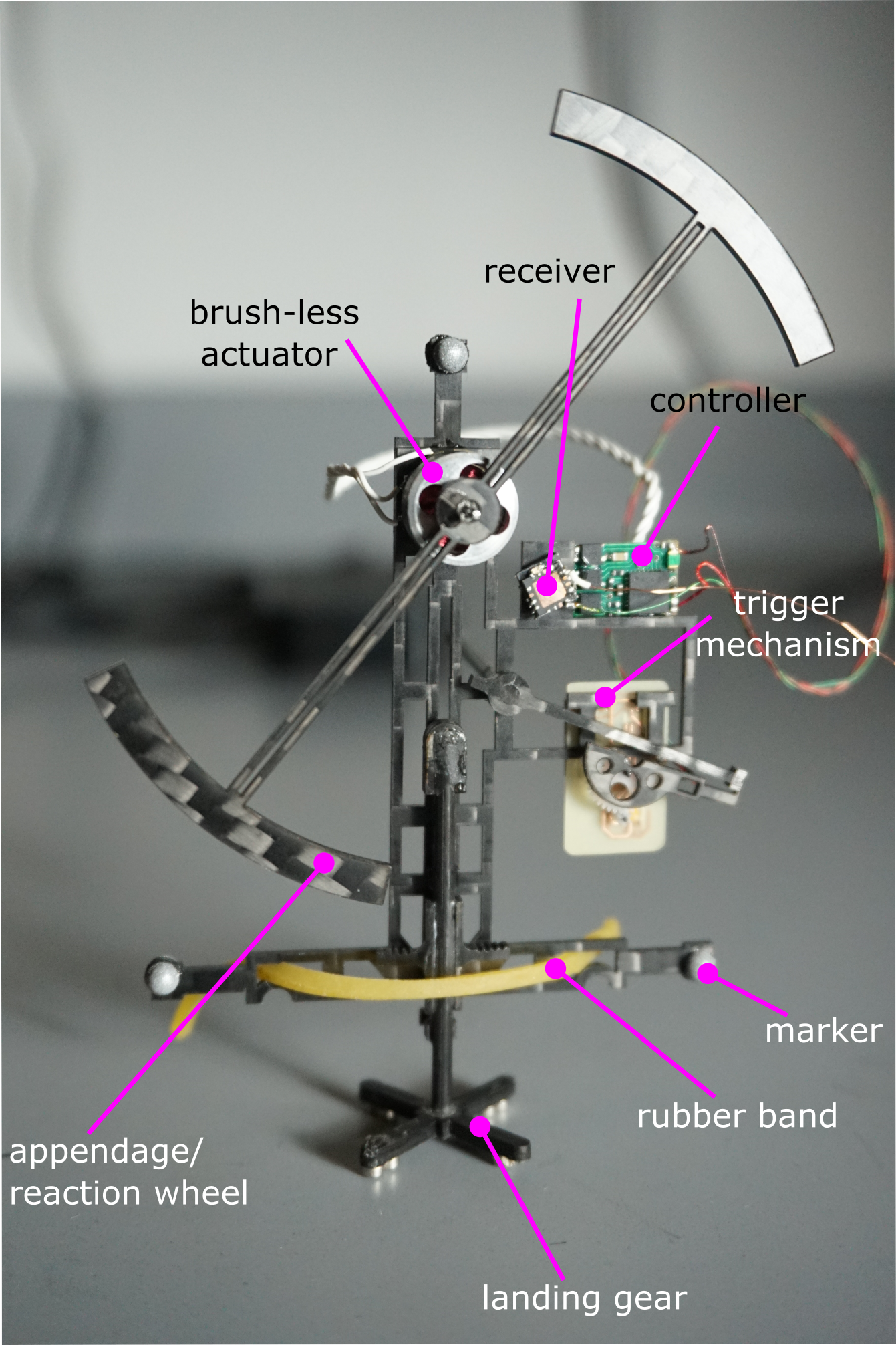}
		\caption{\textit{Harpoon-V.2} developed at Northeastern University to test the idea of inertial dynamics manipulation for fast aerial robot landing on ceilings and steel frames in residential spaces.}
		\label{fig:harpoon}
\end{figure}

Among these maneuvers, landing (or perching), which flying vertebrates do it in one way or another for a variety of reasons (e.g., transition to walking, resting on a perch, hanging from the ceiling of a cave, etc.), is an interesting maneuver to take inspiration from for aerial robot designs. Perching birds rotate their wings so that the aerodynamic drag is increased by creating a high-pressure region inside of the wings and a low-pressure region behind the wings. This brings the wings to a stalled condition at which point the generated lift is equal to zero and the animal falls naturally while employing the legs as a landing gear. 

Bats do it in a radically different way. After the self-created stalled condition, they manifest an acrobatic heels-above-head maneuver that involves catapulting the lower body in a similar way that a free style swimmer flip turns. 

Perching insects and birds have been the source of inspiration and bio-mimicry of them has led to interesting robot designs in recent years. Remarkably, the bio-mimicry of bat-like landing is overlooked mainly because not only the aerodynamics adjustments are involved but also unique design provisions are required to allow for the manipulation of inertial dynamics. 

A part from the ordeals associated with the design and control of a robot that can land similar to a biological bat a number of unique applications can be identified for these systems. For instance, in scenarios wherein maintaining a high vantage point for extended times is demanded (surveillance and reconnaissance) and limited power budget does not allow to hover for extended time periods hanging from elevated structures such as steel frames in buildings can allow these systems not only safely accomplish their missions but also harvest energy within the time period the system is the most vulnerable. 

In addition to the exciting applications, a bat-style landing maneuver is extremely rich in dynamics and control and its characteristics are overlooked. Much of attention has been paid to simpler dynamics such as hovering and straight flight. While mathematical models of insect-style, rotary- and fixed-wing robots of varying size and complexity are relatively well developed, models of airborne, fluidic-based vertebrates locomotion remain largely open due to the complex body articulation involved in their flight. 


The mainstream school of thought inspired by insect flight has conceptualized wing as a mass-less, rigid structure, which is nearly planar and translates -- as a whole or in two-three rigid parts -- through space. In this view, wings possess no inertial effect, are fast that yield two-time-scale dynamics, permit quasi-static external force descriptions, and tractable dynamical system. Unfortunately, these paradigms fail to provide insight into airborne, vertebrate locomotion and an ingredient of a more complete and biologically realistic model is missing, that is, the manipulation of inertial dynamics. The manipulation of inertial dynamics is an under-appreciated aspect in existing paradigms.


The objective of this work is to manifest closed-loop aerial body reorientation and preparation for landing through the manipulation of inertial dynamics using a tiny ballistic robot whose characteristics are carefully scaled to match that of a small bat. It is shown that despite a number of prohibitive restrictions, and lack of multi-thruster designs typically found in quadrotors, extremely fast body reorientations and preparations for landing through inertial dynamics manipulations not only is possible but also is an effective biologically meaningful solution. 

A brief overview of the system will be presented followed by a hybrid-model description of the landing maneuver. A closed-loop controller will be designed and the simulation and experimental results will be reported at the end followed by final remarks and conclusion.

\section{A hybrid-model description of upside-down perching}
A little reflection reveals, however, that a perching bat dynamics like other complex animal behavior, which emerges from complex interactions between neural and sensory-motor systems, can be described with tractable mathematical models. The legitimate question that is posed here is: {\it ''Can a hybrid model capture the dynamics?''} What encourages me to pose the question directly stems from inspecting high-speed images of the landing maneuver of bats and that in a time envelope not exceeding about one fifth of a second before landing completion (touch-down) the appendages are fully retracted strongly suggesting that the aerodynamic forces are less likely to be as important as other forces. Based on a similar observation, \cite{bergou2010fruit} conducted a simulation-based research and concluded that the inertial dynamics manipulation likely are the dominant players in a landing bat.          

It is assumed for two modes: mode (1) with aerodynamics dominance (inertial forces are not negligible but are less effective) and mode (2) within which the inertial dynamics are dominant. In a perching bat, a transition from one mode to another triggered by propreoceptive sensing and neural reflexes occur at a remarkable high speed. 

This hybrid model will allow the individual examination of each mode. Fluid-structure interactions are hard to model, however, the aerial flip turns (backspins) are mainly the result of differential inertial forces and are rigorously explainable by mathematical models. To describe mode (1), the current schemes can take a number of experimentally validated and successful forms \cite{ramezani2017biomimetic}. Therefore, the focus will be mode (2) and that what would be the best strategy to prepare for landing on steel frames upside-down. 


One may raise the concern that how this transition -- i.e., mode (1) to (2) -- will occur in the first place, or, how the aerodynamic forces could vanish quickly? I will draw inspiration from biological examples and that there are a variety of deployable morpho-functional systems to consider \cite{vincent_deployable_2003}. In bats, the transition occur similar to birds by facing the inside of the wings towards incoming flow to initiate stalled conditions at which point the lift force vanishes followed by swiftly collapsing the wings at the onset of the landing maneuver, which almost likely completely zeros the effects of delayed stall and aerodynamic resistance during the backspin motion. 


Without such a mediolateral movement in wings, at high angle of attack maneuvers the aerodynamic forces will be roughly normal to the wing surface at all times resulting in large pressure forces that dominate the shear viscose forces acting parallel to the wings. Because fixed-wing systems have no retraction mechanism in their aerodynamic control surfaces they cannot easily backspin. Rotory-wing systems to be able to preform backspin, they require thruster units that can generate positive and negative aerodynamic force relative to the body and that can be achieved through a number of costly approaches. Another issue concerning fluid-structure interaction can arise here. The ''ceiling effects'' are unknown for us but I believe this phenomena can cause instability in a similar way that ground effects can lead to control issues and instability for near-ground hovering rotory-wing systems. 


An informal comparison of a landing bat and a perching rotorcraft is meaningfully relatable to the comparison between the performance of a pitcher and a robotic arm in throwing the baseball in a baseball game. One employs natural dynamics to perfection to pitch a ball the other is carefully restricted to its predefined joint trajectories completely suppressing the natural dynamics. Of course, unlike the fully-actuated manipulator, a quadrotor is underactuated and its internal dynamics is invariant of the supervisory controller which allows for a limited contribution from the natural dynamics.


\section{Brief Overview of {\it ''Harpoon''} System}
The hypothesis explained earlier is tested with the help of a self-sustained ballistic object called \textit{''Harpoon''} (shown in Fig.~\ref{fig:harpoon}). This robot is considered as the landing gear in a morpho-functional machine. Here, the objective is to reorient the robot towards an \textit{imaginary ceiling surface} and extend (or shoot) the landing-gear towards the ceiling. This scenario is designed after carefully observing the high-speed images of bat landing maneuvers and once successfully reconstructed it will capture the dynamics once inspected by \cite{bergou2010fruit,riskin_bats_2009,bahlman_joseph_w._glide_2013,iriarte-diaz_kinematics_2008}. 

Harpoon platform is laser-cut from a thin carbon fiber plate and hosts: two actuators, two motor controller, a receiver and a processor. The overall design allows for a limited self-sustained sensing, actuation, computation and communication between the robot and master computer. A brush-less motor adjusts the angle of a bob (appendage) with respect to the body. This bob is designed to precisely capture the body to appendage ratio found in bats. A trigger mechanism releases the landing gear once it is faced towards the surface. Since the projectile is charged with massive elastic energy beyond the output performance of the trigger actuator, a mechanism with 1:40 mechanical advantage is designed and fabricated at sub-millimeter-scale resolution as shown in Fig.~\ref{fig:harpoon}.  

The orientation angles including the Euler parameters roll, pitch and yaw are estimated at 200 Hz using an OptiTrack system with six cameras in a 8 m\textsuperscript{3} space and are fed back to a discrete controller with known asymptotic stability properties. Then, the computed control inputs are fed back to the robot in a wireless communication bus in Ethernet frames. This architecture allows for a robust communication for fast sensing and actuation. The computations take place off-line and it allows for an expensive real-time processing otherwise impossible with the limited on-board processing capabilities.

Additionally, a computer-automated mechanism, called \textit{releaser}, ensures the robot is released with zero-angular momentum, therefore, the ballistic motion will be of zero-angular-momentum nature \cite{arabyan_distributed_nodate}. This condition may put restrictions on the boundaries of the trajectories from mode (1), however, a change in the mechanism design can easily provide non-zero-angular-momentum turns, which can easily allow for the study of non-zero-angular-momentum backspins. To completely suppress the aerodynamic effects, wings are excluded from the model and the host morpho-functional machine will be unveiled in the future works.

\section{Manipulation of inertial dynamics}
The system dynamics and the equations of motion are described in a coordinate-free fashion, i.e., a system on manifold without considering any local coordinate charts. The kinetic energy is considered as a Riemannian metric and in writing the discrete Hamilton's principle associated Riemannian inter-connections are considered. The discrete-time model on manifold with body-fixed forces and invariant kinetic energies is erected based on the Lie group variational integrator evolving on $SE(3)\times SO(3)$ which unlike a rigid body results in a nontrivial dynamical model. 

The numerical integrator obtained from the discrete variational principle exhibits excellent geometric conservation properties \cite{hairer2006geometric} and because of this it is used here. This means that the robot attitude automatically evolves on the rotation group embedded in the space of special orthogonal matrices whereas the required angular velocity of appendage is computed at the level of the Lie algebra and the matrix exponential are employed to update the reference solutions. 

Last, an almost-globally stabilizing, geometric discrete-time controller based on \cite{sanyal2009inertia} will be applied to regulate the attitude in preparation for landing.

\subsection{Notation and Assumptions}
\begin{figure}[!t]
	\centering
		\includegraphics[width=0.3\textwidth]{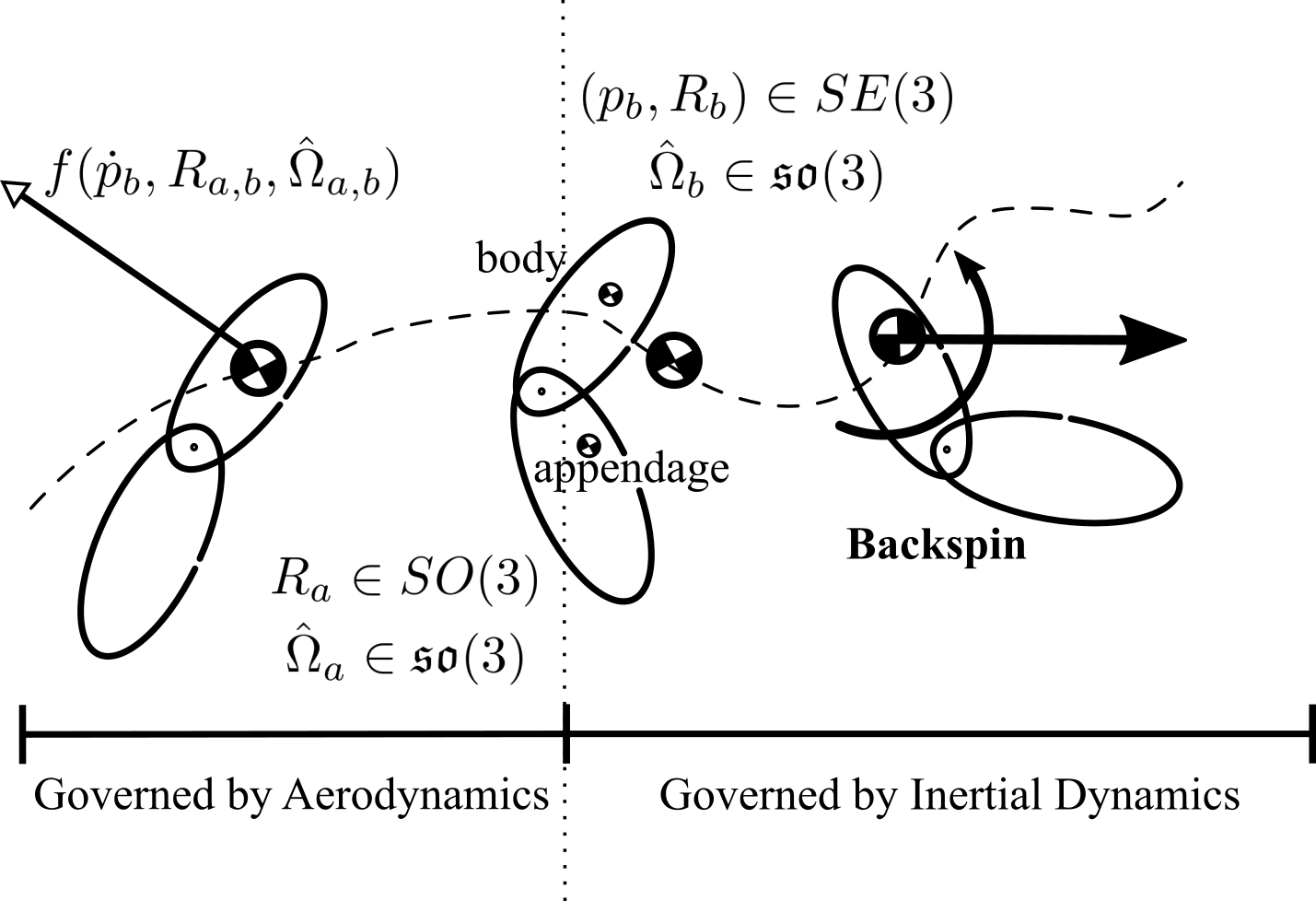}
		\caption{A hybrid model is considered to explain the landing maneuver and Moser-Veselov description of the discrete-time inertial manipulation is considered on $SE(3)\times SO(3)$.}
		\label{fig:hybridmodel}
\end{figure}

It is assumed that a body coordinate frame is fixated to the robot and it position, orientation and angular velocity are denoted by $p$, $R_b$ and $\Omega_b$, respectively. Another body coordinate frame is attached to the appendage where the orientation and angular velocities are denoted by $R_a$ and $\Omega_a$, respectively. Therefore, the configuration space (C-space) is $\left\{p,R_b,R_a,\Omega_b,\Omega_a\right\}$. The physical properties including the inertia matrix and mass are denoted by $J_{a,b}$ and $m_{a,b}$. A small change in $R_i$ is denoted by $F_i$ and $\eta_i$ denotes variations on $SO(3)$. The Lagrangian functional, kinetic and potential energy are denoted by $L$, $K$ and $V$, respectively. Other widely used notations such as $\hat{(.)}$ for wedge operator, $X[k]$ for a discrete value of $X$ at the kth sample, $\mathrm{tr}(.)$ for the trace of a matrix, $(.)^T$ as the transpose operator, $\mathrm{eig}(.)$ as the eigen-values of a matrix and $\delta(.)$ as the variation operator are adopted. The control input is denoted by $u$, the identity matrix is denoted by $I$, $e_3$ is the z-axis unit vector and $r_{a,b}$ are the distance between the joint and the center of mass in body and appendage.       


\subsection{Moser-Veselov Description of the Discrete-time, Constrained Dynamics}



Sensing and actuation occur in discrete-time, therefore, the discrete-time version of the dynamical system on $SE(3)\times SO(3)$ is considered and that arises the constrained dynamics once investigated by \cite{moser1991discrete}. Indirect methods based on Rodrigues' approach to approximate matrix exponential and first-order approximations of variations on Lie groups are considered as well which will be explained in this section.

\subsubsection{Schur-form update law to $R_b[k]$ and $R_a[k]$} The discretized version of Hamilton's principle determines the equations of motion from a variational problem. The Lie group variational dynamics for the system of body and appendage when external forces are absent are given by:
\begin{align}
    h\hat{\Pi}_b[k]&=F_b[k]J_b-J_b F_b^T[k], \label{eq:moser-veselov-1}\\
    R_b[k+1]&=R_b[k]F_b[k], \label{eq:moser-veselov-2}\\
    \Pi_b[k+1]&=F_b^T[k]\Pi_b[k]+hu, \label{eq:moser-veselov-3}\\
    0&\leq \left(h\hat{\Pi}_b[k]\right)^2+4\left(\frac{1}{2}\mathrm{tr}(J_b)I-J_b\right)^2 \label{eq:moser-veselov-4}
\end{align}
\noindent where $u\in\mathbb{R}^3$ allows for the inertial dynamics contributions from the appendages. In Eqs.~\ref{eq:moser-veselov-1}, \ref{eq:moser-veselov-2}, \ref{eq:moser-veselov-3} and \ref{eq:moser-veselov-4}, the solution to the discrete-time dynamical system can be resolved first by obtaining an answer to $F_b[k]$, which involves solving the continuous-time algebraic Riccati equation (ARE) given by Eq.~\ref{eq:moser-veselov-1}. However, the solutions to the ARE are not unique and for these solution to belong to special orthogonal solutions ($F_b[k]\in SO(3)$) the convex constraint given by Eq.~\ref{eq:moser-veselov-4} must be satisfied \cite{moser1991discrete}. In general, a change in the orientation of the body $F_b[k]$ cannot be uniquely mapped to $\Omega_b$ as $F_b[k]$ are minuscule changes in $R_b[k]\in SO(3)$. To obtain $F_b[k]$, the steps from \cite{bunse1986symplectic} are applied. The Hamiltonian matrix is given by
\begin{equation}
H[k]=\left[
\begin{array}{cc}
\frac{h\hat{\Pi}_b[k]}{2}&I\\
\frac{\left(h\hat{\Pi}_b[k]\right)^2}{4}+J_b^2&\frac{h\hat{\Pi}_b[k]}{2}
\end{array}
\right]
\end{equation}
\noindent and after synthesizing the Schur form of the Hamiltonian Matrix $\gamma[k]^T H[k]\gamma[k]=\left[\begin{array}{cc}
     \alpha_1&\alpha_2  \\
     0&\alpha_3 
\end{array}\right]_{\mathrm{Schur},k}$, $\gamma[k]$ is partitioned as $\gamma[k]=\left[\begin{array}{cc}
     \gamma_1&\gamma_2  \\
     \gamma_3&\gamma_4 
\end{array}\right]_k$. Hence, the special orthogonal solutions to the ARE are given by 
\begin{equation}
    F_b[k]=\left(\gamma_3\gamma_1^{-1}+\frac{h\hat{\Pi}_b[k]}{2}\right)J_b^{-1}.
\end{equation}

\subsubsection{Rodrigues' approximation to the variations in $R_b[k]$ and $R_a[k]$} In addition to the Schur-form solutions, another approach is considered. In this way, special orthogonal solutions to Eqs.~\ref{eq:moser-veselov-1}-\ref{eq:moser-veselov-4} are iteratively resolved by expressing $F_i[k]\in SO(3),~~i\in{a,b}$ as an exponential form of $\hat{f}_i[k]\in\mathfrak{so}(3)$ where $f_i[k]$ belongs to $\mathbb{R}^3$. The Rodrigues' formula for the changes in the body orientation gives
\begin{align*}
    F_b[k]&=I+\frac{\sin\|f_b[k]\|}{\|f_b[k]\|}\hat{f}_b[k]\\
    &+\frac{1-\cos\|f_b[k]\|}{\|f_b[k]\|^2}\left(\hat{f}_b[k]\right)^2
\end{align*}
\noindent and the ARE given in Eq.~\ref{eq:moser-veselov-1} is re-written into the equivalent vector form given by

\begin{align*}
    h\Pi_b[k]&=\frac{\sin\|f_b[k]\|}{\|f_b[k]\|}J_bf_b[k]\\
    &+\frac{1-\cos\|f_b[k]\|}{\|f_b[k]\|^2}\hat{f}_b[k]J_bf_b[k]
\end{align*}

Rodrigues' approach was considered in addition to the Schur method to provision for the numeric difficulties endured when attempted to use Eqs.~\ref{eq:moser-veselov-1}-\ref{eq:moser-veselov-4} in a real-time reference trajectory governor. The Eqs.~\ref{eq:moser-veselov-1}-\ref{eq:moser-veselov-4} yield the symplectic geometry of the problem in hand and is followed by computational issues about solutions in finding invariant subspaces of a Hamiltonian matrics. Other methods that could take into account the particular structure of Hamiltonian matrices are not less costly. 
Worthy of noting is that it was noticed -- in this experimentally motivated work -- extra care must be paid to as not all of the sampling rates (time intervals denoted by $h$) can guarantee the existence and uniqueness of the solutions in Eqs.~\ref{eq:moser-veselov-1}-\ref{eq:moser-veselov-4}. Intuitively, this makes sense as a relatively large time interval leads to an approximations of $F_i[k]$ that violates the topological structure of $SO(3)$. The following equation can help observe this problem  
\begin{align}
F_i[k]  = h\hat{\Omega}_i[k] + I \label{eq:F-approx}
\end{align}
\noindent Obviously, $h$ cannot take any arbitrary number. This equation is the direct result of defining $F_i[k]$ such that $R_i[k+1]=R_i[k]F_i[k]$. Using the kinematic relationship and considering an approximation for $\hat{\Omega}_i[k]\approx R^T_i[k]\left(\frac{R_i[k+1]-R_i[k]}{h}\right)$ Eq.~\ref{eq:F-approx} is obtained.

Almost all of these methods resolve unique positive definite solutions for ARE when sample intervals are small, however, showing how the size of the sample intervals can violate the Hamiltonian matrix with no imaginary eigenvalues is well beyond the scope of this work. 

Although the sufficient and necessary convex quadratic constraint in Eq.~\ref{eq:moser-veselov-4} is a strong condition that secures the existence of such a solution, its dependence to the sampling time intervals can be limiting. 

When $\left(h\hat{\Pi}_b[k]\right)^2/4+J_b^2$ is positive definite then Eq.~\ref{eq:moser-veselov-1} has a unique positive semi-definite solution $S[k]\geq 0$, i.e., $\mathrm{eig}(S[k]+\hat{\Pi}_b[k]/2)$ has positive real parts. Consequently, $F_b[k]=(S[k]+h\hat{\Pi}_b[k]/2)J_b^{-1}$ is the special orthogonal matrix being sought. When $\left(h\hat{\Pi}_b[k]\right)^2/4+J_b^2$ is positive definite an invertible matrix $E[k]$ exists such that $\left(h\hat{\Pi}_b[k]\right)^2/4+J_b^2=E[k]^TE[k]$. Consequently, $(h\hat{\Pi}_b[k]/2,I)$ and $(E[k],h\hat{\Pi}_b[k]/2)$ will be stabilizable and detectable, respectively, which results in $\mathrm{eig}(S[k]+h\hat{\Pi}_b[k]/2)$ with positive real parts. Since $\mathrm{eig}(\pm h\hat{\Pi}_b[k]/2-I)$ are the form $-1\pm\beta_i$ they have negative real part. Therefore, $(h\hat{\Pi}_b[k]/2+I)$ and $(-h\hat{\Pi}_b[k]/2,E^T)$ are stabilizable, that is, $(h\hat{\Pi}_b[k]/2,I)$ is stabilizable and $(E,h\hat{\Pi}_b[k]/2)$ is detectable.  

\subsubsection{Base-line coordinate-free system on $SO(3)$ manifold} Based on Hamilton's Principle, the action integral is given from the Lagrangian $L=K-V$ and its variations are zero. Because the configuration group is the Lie group, this variation should be consistent with this geometry, therefore, the varied $\Omega_i,i\in{a,b}$ in continuous-time mode is given by
\begin{equation}
\delta \Omega_i = \hat{\Omega}_i\eta_i + \dot{\eta}_i,~i\in{a,b} \nonumber \\
\end{equation}
\noindent and in discrete-time mode is given by
\begin{equation}
\delta \Omega_i[k] = \frac {1} {h} \left(F_i[k] \eta_i[k+1] - \eta_i[k]\right),~i\in{a,b}. \nonumber\\
\end{equation}

Consequently the variations of the action integral are obtained. In driving the variations of continuous-time action integral, ${\Omega_i^T} {J_i} {\dot {\eta}_i}  = {\frac {d}{dt}} ({\Omega_i^T} {J_i} {\eta_i}) - {\frac {d {\Omega_i^T}} {dt}} {J_i} {\eta_i}$ is used. Note that the continuous- and discrete-time variations vanish at the time boundaries, that is, $t_0$ and $t_f$ for the continuous system and $k=0$ and $k=N$ for the discrete system. From Hamilton's principle, the above continuous- and discrete-time variations in the action integral should be zero for all variation of Lie algebra. Hence, the continuous time system on manifold is given by:
\begin{equation}
    \dot{x} = G_1(x)+G_2(x)u
\end{equation}
\noindent where $x=(p;R_b;R_a;\dot{p};\Omega_b;\Omega_a)\in SE(3)\times SO(3)$ and $u\in \mathbb{R}^3$ are the states and control inputs, respectively; $G_1(x)=\psi^{-1}(x)\phi(x)$ and $G_2(x)=\psi^{-1}(x)\left[0_{3\times 3},I_{3\times 3},-R_a^TR_b\right]^T$ are obtained after sorting the varaitions and applying some basic algebra. The symmetric matrix $\psi(x)\in\mathbb{R}^{9\times 9}$ is given by
\begin{equation}
\psi(x)=
\left[
\begin{array}{ccc}
     -MI&\xi_b\nu_b&-\xi_a\nu_a\\
     \xi_b\nu^T_b&r_b\xi_b\hat{e}^2_3-J_b&r_a\xi_b\nu^{T}_b\nu_a\\
     -\xi_a\nu^T_a&r_a\xi_b\nu_a^T\nu_b&r_a\xi_a\hat{e}^2_3-J_a
\end{array}
\right]
\end{equation}
\noindent and $\phi(x)\in\mathbb{R}^{9}$ is given by
\begin{equation}
\phi(x)=
\left[
\begin{array}{l}
    \xi_bR_b\kappa_b-\xi_aR_a\kappa_a-Mge_3 \vspace{0.25cm}\\ 
    \hat{\Omega}_b\Pi_b+2\xi_b\zeta^T_b\dot{p}+r_b\xi_b\hat{e}_3\kappa_b+\dots\\
    2r_b\xi_a\zeta^T_b\nu_a\Omega_a+\xi_bg\hat{\nu_b}e_3-r_b\xi_a\nu_b\zeta_a\Omega_a \vspace{0.25cm}\\ 
    \hat{\Omega}_a\Pi_a-2\xi_a\zeta^T_a\dot{p}+r_a\xi_a\hat{e}_3\kappa_b-\dots\\
    2r_a\xi_b\zeta^T_a\nu_b\Omega_b-\xi_bg\hat{\nu_a}e_3-r_a\xi_b\nu_a\zeta_a\Omega_b
\end{array}
\right]
\end{equation}
\noindent where $M=\sum_i m_i$, $\xi_{i}=m_ar_i$, $\nu_i=R_i\hat{e}_3$, $\zeta_i=R_i\hat{\Omega}_i\hat{e}_3$ and $\kappa_i=\hat{\Omega}^2_ie_3$ for $i\in{a,b}$. The discreteization of this continuous dynamics (Eq.\ref{eq:moser-veselov-1}-\ref{eq:moser-veselov-4}) is achieved by applying the discrete-time variation explained above and applying Hamilton's principle. Next, I will briefly describe the closed-loop feedback used to regulate the body orientation. 

\subsection{Discrete feedback control}
A scalar measure of the attitude error is given by the rotation angle $\theta[k]$ about the eigen-axis needed to rotate the robot form $R_b[k]$ to the desired attitude $R_d[k]$ and is given by $\theta[k]=\cos^{-1}\left(\frac{1}{2}\left[\mathrm{tr}R^d_T[k]R_R[k]-1\right]\right)$. This metric is applied to indicate the time of launching the landing gear. 

The controller applied here does not require any knowledge in regard to the mass and inertia of the robot body and appendage. A major limitation in designing the closed-loop feedback for this constrained dynamics is that not all of the arbitrary reference trajectories can be enforced by the tracking controller and the tracking problem is subject to the constraints given by Eq.~\ref{eq:moser-veselov-4}. 

The issue of constraint satisfaction in nonlinear systems affine- and non-affine-in-control and its separation from the issue of closed-loop system design have been addressed very extensively. Almost all of these works do so by introducing a reference governor, an auxiliary nonlinear device that operates between the reference command and the input to the closed-loop system. A similar approach to \cite{gilbert2002nonlinear} is adopted here. 

The control input with known stability properties \cite{sanyal2009inertia} is given by
\begin{align}
u = K_1 \Pi_{b}[k] - e[k]
\end{align}
\noindent where $K_1$ is the feedback gain and $e[k]$ can be calculated by the following relation,
\begin{align}
\hat e[k] = K_2 \Gamma^T[k]R_b[k] - R_b^T[k]\Gamma[k]K_2 
\end{align} 
\noindent where $K_2$ is a symmetric positive-definite matrix and $\Gamma[k]$ is the reference input to the body orientation and is resolved by finding $\alpha[k]$ in a nonlinear optimizer subject to the following constraints:
\begin{align}
0&\leq \alpha[k], \vspace{0.25cm}\\
\alpha[k]&\leq 1, \vspace{0.25cm}\\
0&\leq \left(h\hat{\Pi}_b[k]\right)^2+4\left(\frac{1}{2}\mathrm{tr}(J_b)I-J_b\right)^2,\vspace{0.25cm}\\
0&=\Gamma[k] - \left(R_d[k] \Gamma^T[k-1]\right)^{\alpha[k]} \Gamma[k-1].
\end{align}
\noindent Note that $R_d[k]$ is the desired body orientation before considering the convex constraint given by Eq.~\ref{eq:moser-veselov-4}.
\section {Numerical \& Experimental Results}
The appendage inertial model is scaled to account for 16\% of the total body mass according to \cite{iriarte-diaz_kinematics_2008,iriarte-diaz_whole-body_2011}. A total estimated body weight of 30 gr is considered, which is obtained after assuming a density of $2\times 10^3 kg.m^{-3}$ according to \cite{iriarte-diaz_whole-body_2011}, while equally distanced bobs from the body joint (shown in Fig.~\ref{fig:hybridmodel}) are considered. A simple body geometry is considered for a seamless inertia matching.

When only the dynamics restricted to the longitudinal plan is considered the C-space is the two-sphere $S^2=\left\{q\in\mathbb{R}^3|q.q=1\right\}$, where $q$ is a unit vector (Note that the body and appendage angles are embedded in $q$). The tangent space at $q$, namely $T_qS^2$, is identified with $T_qS^2\approx\left\{\omega\in\mathbb{R}^3|q.\omega=0\right\}$ where $\omega\in\mathbb{R}^3$ represents the angular velocity. 

\begin{figure}[!h]
\begin{subfigure}{.5\textwidth}
  \centering
  \includegraphics[width=.25\linewidth]{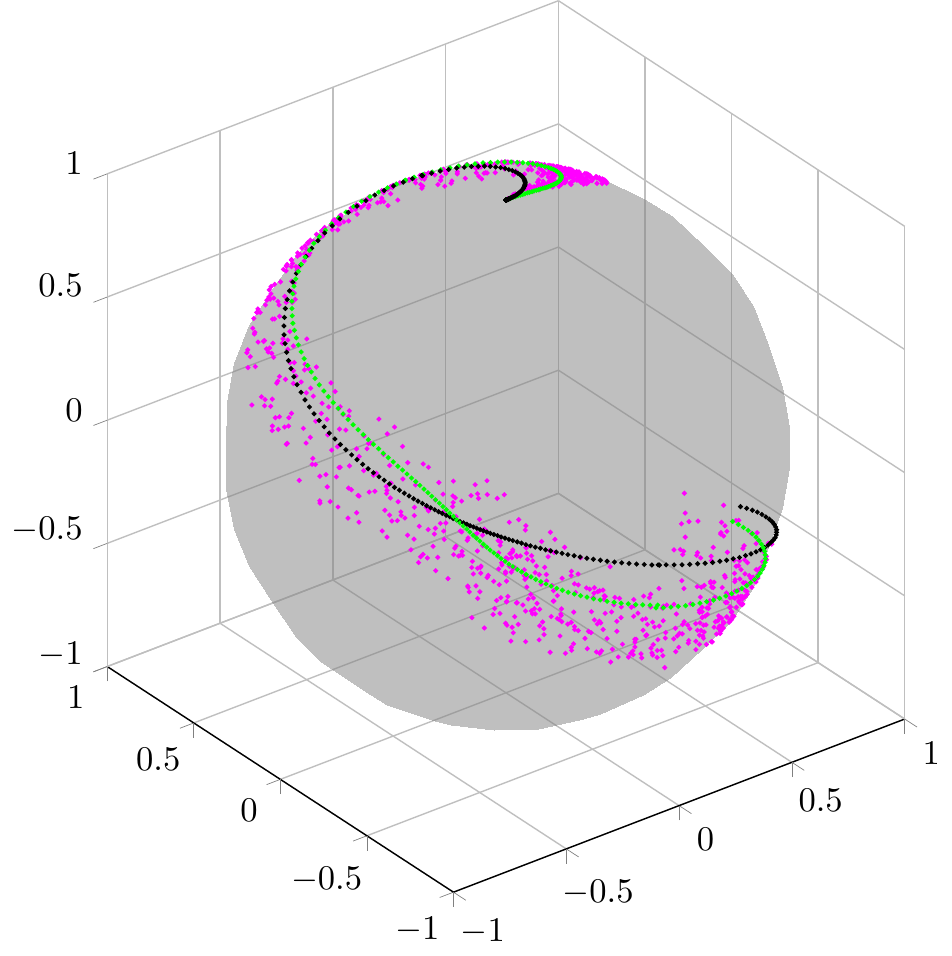}
  \caption{Resolved reference trajectory (green) reaching to 120 deg subject to the constraints in Eq.~\ref{eq:moser-veselov-4} shown on a two-sphere}
\label{fig:RGS2-a}
\end{subfigure}

\begin{subfigure}{.5\textwidth}
  \centering
  \includegraphics[width=.25\linewidth]{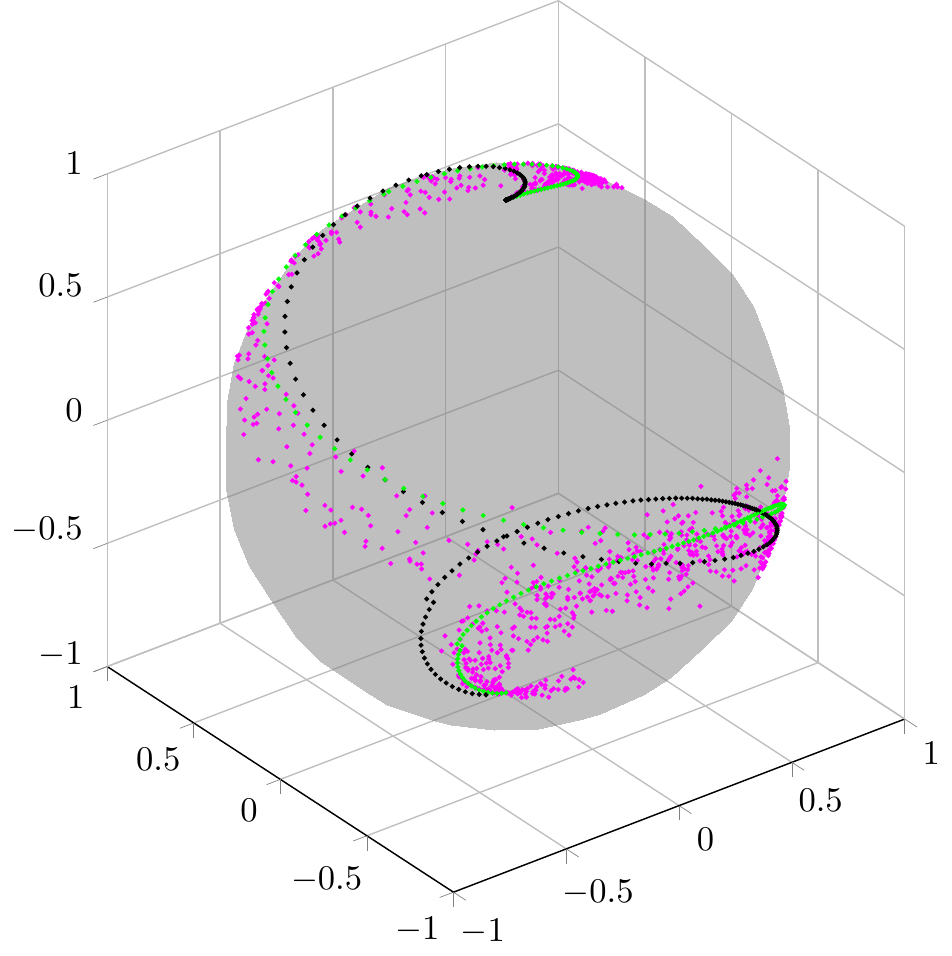}
  \caption{Resolved reference trajectory reaching to 180 deg}
  \label{fig:RGS2-b}
\end{subfigure}
\end{figure}

\begin{figure}
\begin{subfigure}{.5\textwidth}
  \centering
  \includegraphics[width=.5\linewidth]{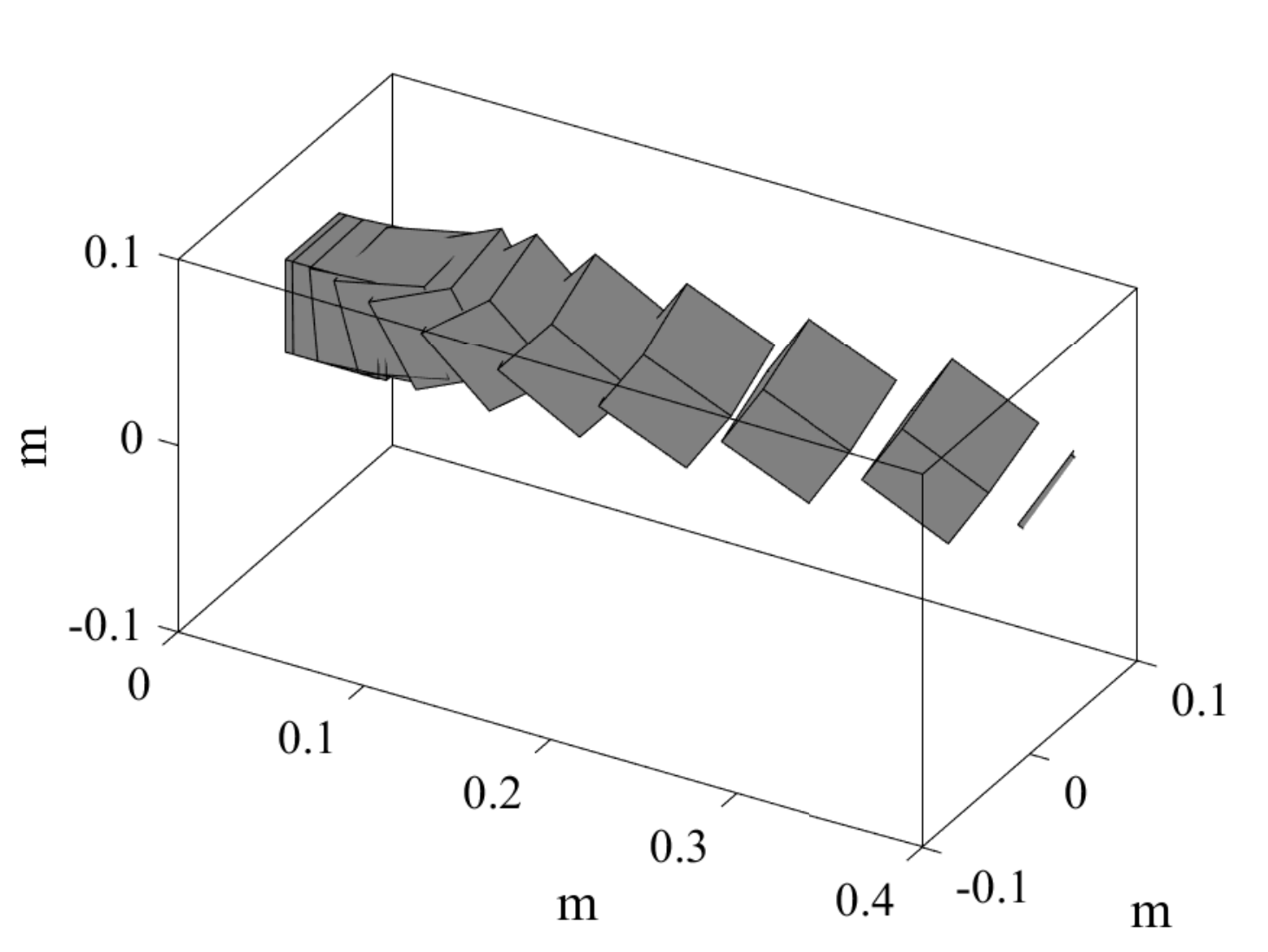}
  \caption{Simulated body motion }
  \label{fig:rigidbody-a}
\end{subfigure}\\
\begin{subfigure}{.5\textwidth}
  \centering
  \includegraphics[width=.5\linewidth]{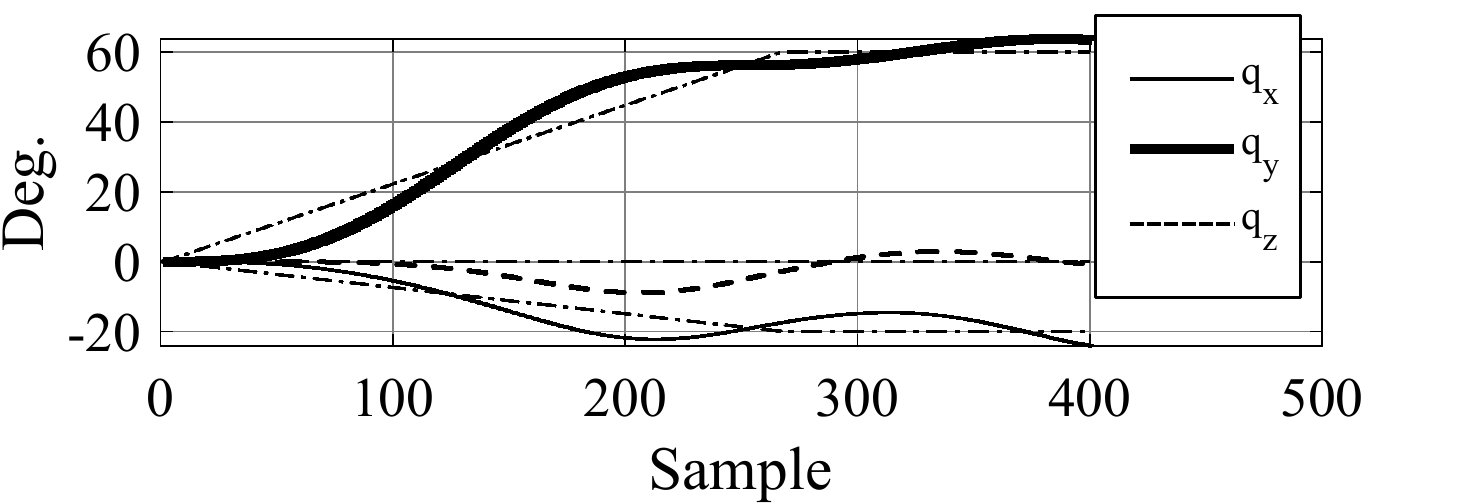}
  \caption{Simulated appendage movements to achieve $-20$ and $60$ degrees variations in the roll and pitch angles}
  \label{fig:rigidbody-b}
\end{subfigure}
\end{figure}

A fixed desired body orientation $R_d$ is constructed according to the Euler convention by ramping the pitch angles to 120 and 180 deg and updated by the reference governor subject to the constraints explained above (shown in Fig.~\ref{fig:RGS2-a} and \ref{fig:RGS2-b}). Samples (purple dots) are considered by the optimizer at the neighborhood of the desired angles (black dots) before resolving the actual trajectories. Note that due to issues in the cluttered raster image some of the purple dots are plotted. In Fig.~\ref{fig:rigidbody-a}, the orientations of a 3D-rigid body representing the body (note that the appendage is not shown) are shown along the time axis for a simultaneous tracking of roll and pitch angels  -20 and 60 deg, respectively. In Fig.~\ref{fig:rigidbody-b}, the simulated Euler angles are shown. 

\begin{figure}[!t]
	\centering
		\includegraphics[width=0.5\textwidth]{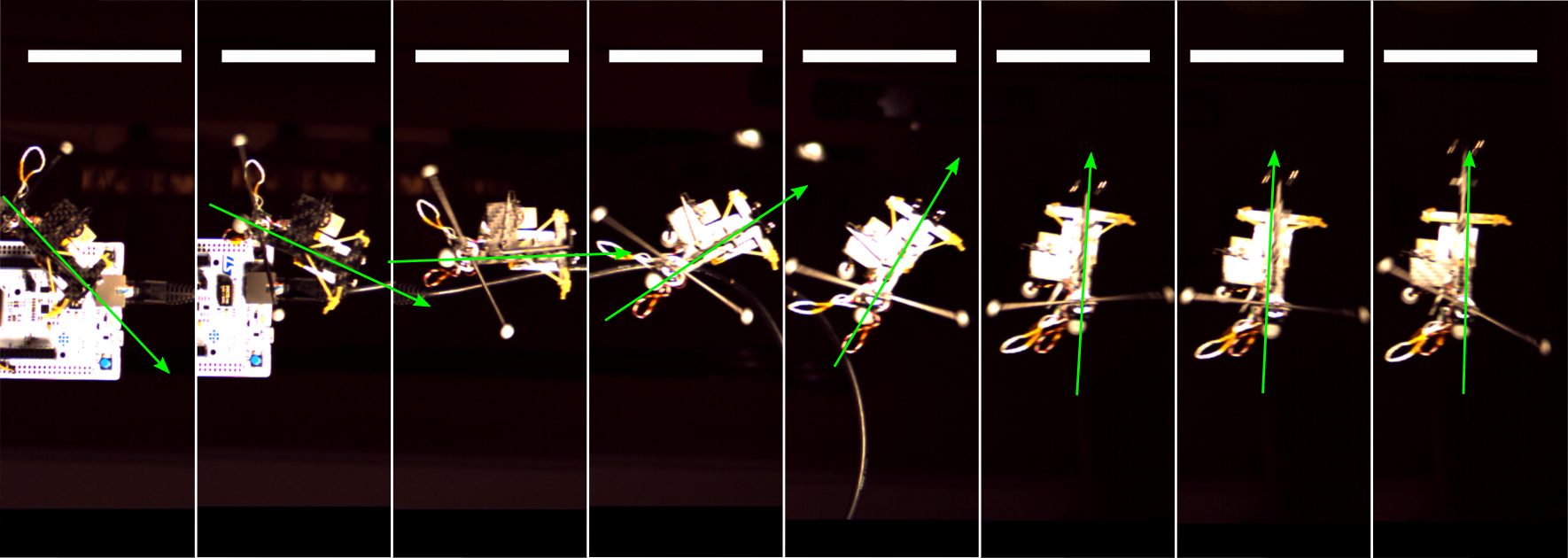}
		\caption{Snapshots of Harpoon while reorienting towards the {\it virtual ceiling} to launch the landing gear}
		\label{fig:snapshots}
\end{figure}

\begin{figure}
\centering
\includegraphics[width=.5\linewidth]{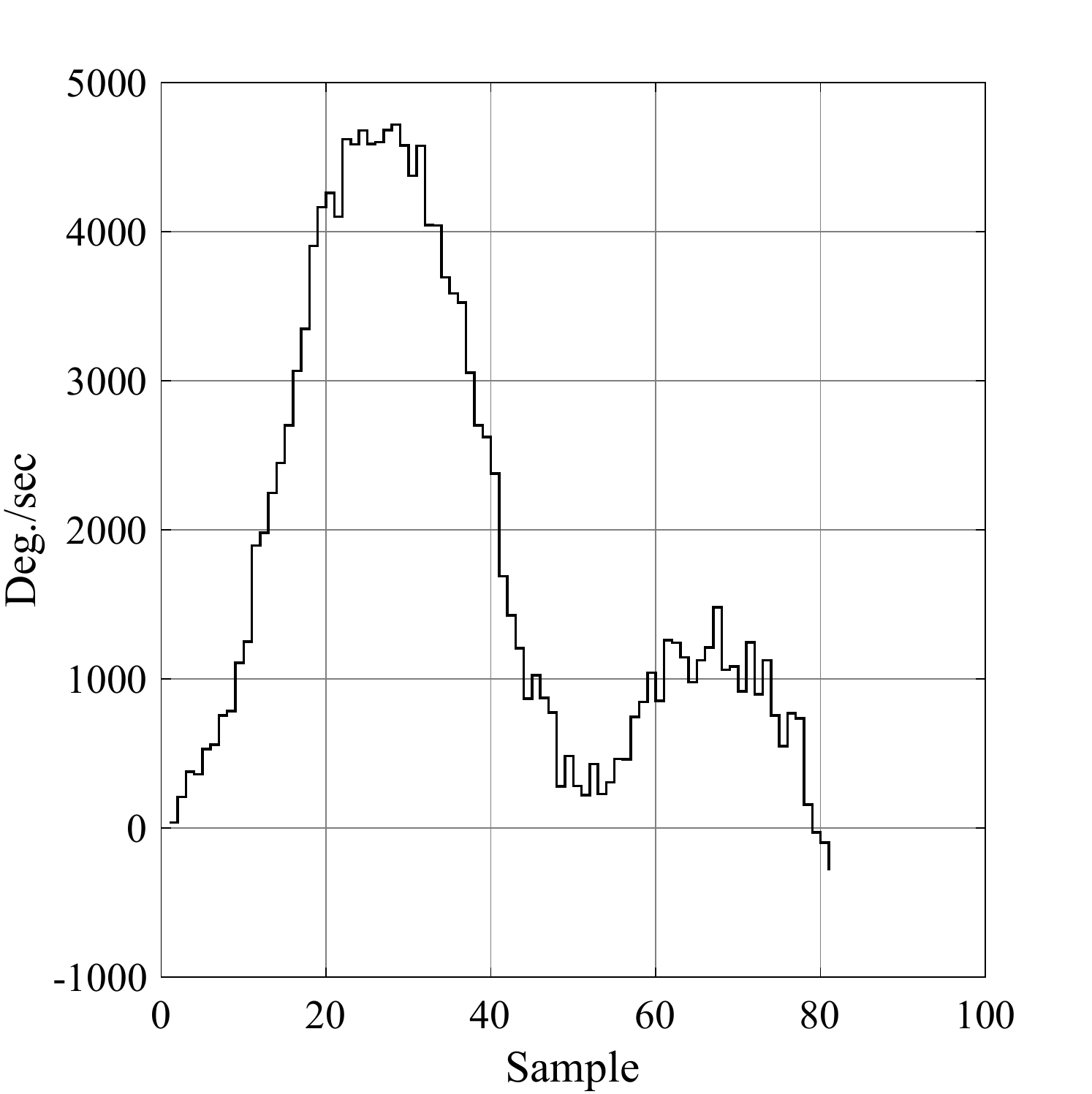}
\caption{Appendage angular velocity $\Omega_{ay}$}
\label{fig:appangularate}
\end{figure}

 In experiment, only the longitudinal motion is considered. The framework was tested in a series of experiments devised with an off-board computer, motion capture system and high-speed imaging equipment. In Fig.~\ref{fig:snapshots}, the snapshots from one experiment are shown. The computer-automated releaser mechanism releases the robot with zero-angular momentum in a free fall configuration while the motion capture system captures the body and appendage orientations. The commanded $\Omega_a$ (shown in Fig.~\ref{fig:appangularate}) are resolved and transmitted to the robot for the desired body orientation angle 180 deg. 

\section {Conclusion}
The ability of animals to reorient their bodies mid-air without the benefit of net external torques is known yet remarkably has been overlooked in aerial drone design. Instead, current designs consider thrust vectoring for agile aerial robotics, which is known to be limited based on aerodynamics laws. This work used a tiny robot called \textit{Harpoon} with prohibitive restrictions associated with size and payload not allowing for multi-thruster designs typically found in quadrotors and demonstrated that extremely fast body reorientation and preparation for upside-down landing unique to bats is possible through the closed-loop manipulations of inertial dynamics in the robot. To prepare for landing, a rubber-band-propelled landing gear design and associated trigger mechanism was proposed. The closed-loop manipulations of inertial dynamics took place based on a symplectic description of the dynamical system (body and appendage), which is known to exhibit an excellent geometric conservation properties.

\bibliographystyle{IEEEtran}
\bibliography{ref}
%
 
\end{document}